\documentclass[10pt,twocolumn,letterpaper]{article}

\usepackage{iccv}
\usepackage{times}
\usepackage{epsfig}
\usepackage{graphicx}
\usepackage{amsmath}
\usepackage{amssymb}
\usepackage{booktabs}
\usepackage{colortbl}
\usepackage[dvipsnames,svgnames]{xcolor}
\usepackage{multirow}
\usepackage{xcolor}
\usepackage{subcaption}
\definecolor{mycolor}{RGB}{255, 105, 180}

\usepackage[breaklinks=true,bookmarks=false]{hyperref}

\iccvfinalcopy 

\def\iccvPaperID{****} 

\ificcvfinal\fi

\begin{document}

\title{Box-DETR: Understanding and Boxing Conditional Spatial Queries}

\author{Wenze Liu\\
\and
Hao Lu\thanks{Corresponding author}\\
\and
Yuliang Liu\\
\and
Zhiguo Cao\\
\and
School of Artificial Intelligence and Automation, \\
Huazhong University of Science and Technology, China\\
{\tt\small \{wzliu,hlu,ylliu,zgcao\}@hust.edu.cn}
}

\maketitle
\ificcvfinal\thispagestyle{empty}\fi

\begin{abstract}
   Conditional spatial queries are recently introduced into DEtection TRansformer (DETR) to accelerate convergence. In DAB-DETR~\cite{liu2021dab}, such queries are modulated by the so-called conditional linear projection at each decoder stage, aiming to search for positions of interest such as the four extremities of the box. Each decoder stage progressively updates the box by predicting the anchor box offsets, while in cross-attention only the box center is informed as the reference point. The use of only box center, however, leaves the width and height of the previous box unknown to the current stage, which hinders accurate prediction of offsets. We argue that the explicit use of the entire box information in cross-attention matters. In this work, we propose Box Agent to condense the box into head-specific agent points. By replacing the box center with the agent point as the reference point in each head, the conditional cross-attention can search for positions from a more reasonable starting point by considering the full scope of the previous box, rather than always from the previous box center. This significantly reduces the burden of the conditional linear projection. Experimental results show that the box agent leads to not only faster convergence but also improved detection performance, e.g., our single-scale model achieves $44.2$ AP with ResNet-50 based on DAB-DETR. Our Box Agent requires minor modifications to the code and has negligible computational workload. Code is available at \url{https://github.com/tiny-smart/box-detr}.
\end{abstract}

\section{Introduction}

\label{sec:intro}
\begin{figure}[!t]
	\centering
	\includegraphics[width=\linewidth]{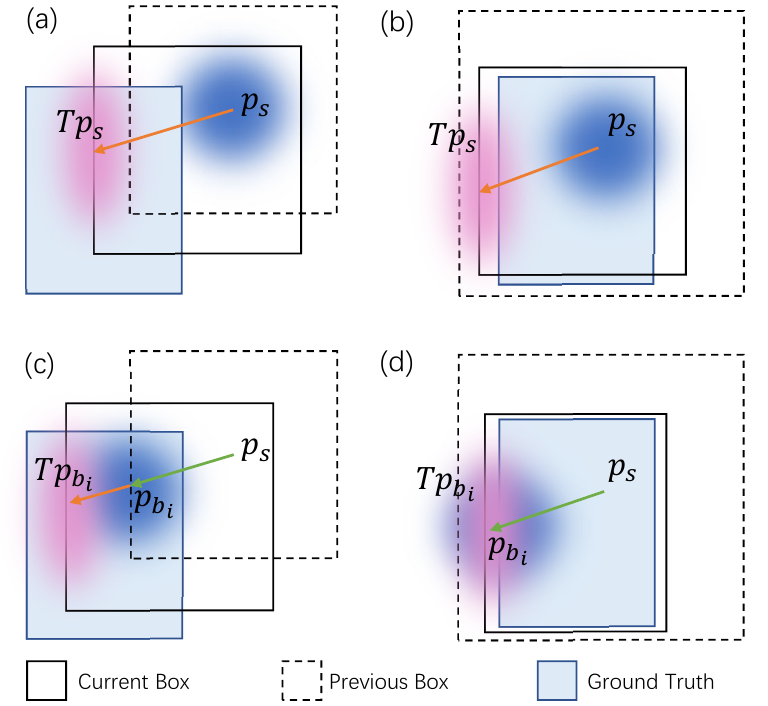}
	\caption{\textbf{Two examples on the function of conditional linear projection $\mathbf{T}$ in DAB-DETR (top) and Box-DETR (bottom) in the $i$-th cross-attention head 
    specific to the left side of the box.} 
    In DAB-DETR, the previous box center is used as the reference point $\mathbf{s}$. Its sine embedding $\mathbf{p}_s$ marks a Gaussian-like contour centered at $\mathbf{s}$ by inner product with the position embedding of the key. $\mathbf{T}$ 
    moves 
    $\mathbf{p}_s$ to 
    search for the current left side by $\mathbf{T}\mathbf{p}_s$. Box-DETR seeks a new reference point $\mathbf{b}_i$ within the previous box beforehand and then uses $\mathbf{T}$ to find the left side of the object given $\mathbf{p}_{b_i}$. If the previous box already covers the object box, then the new reference point can be right on the left side of the object box. 
	}
	\label{fig:T_comparison}
\end{figure}


DEtection TRansformer (DETR)~\cite{carion2020end} has recently posed 
object detection as a set prediction problem, eliminates the need of anchor boxes and Non-Maximum Suppression (NMS) post-processing, and thus establishes a fully end-to-end detection framework. 
Yet, it suffers from slow training convergence and requires up to $500$ training epochs.

To alleviate 
the convergence problem, a stream of work~\cite{zhu2020deformable,gao2021fast,meng2021conditional,liu2021dab,li2022dn,zhang2022dino,jia2022detrs} 
has proposed different solutions. At the model level, 
the key idea is to reuse the decoder embedding generated by the previous decoder stage and constrain the query to focus on specific spatial regions instead of the entire feature map. 
In particular, Deformable DETR~\cite{zhu2020deformable} 
uses the query and the reference point to generate positions of interest and 
implements sparse, deformable attention on a small set of keys. SMCA~\cite{gao2021fast} uses position-related Gaussian heat maps to modulate the attention behavior. To improve the quality of the spatial query, Conditional DETR~\cite{meng2021conditional} generates the conditional spatial query from the decoder embedding and the reference point. In this way, the conditional cross-attention accomplishes two jobs: i) searching for positions of interest starting from the reference point, and ii) expand and change the attention shape to cover representative feature points (Fig.~\ref{fig:T_comparison}(a) (b)). 
As the follow-up work of Conditional DETR, DAB-DETR~\cite{liu2021dab} further adopts the anchor boxes as queries, and each decoder stage updates the box query layer-by-layer by predicting the box offsets. However, while the sparse attention in Deformable DETR can directly sample points from the previous box to utilize the box scale (width and height) information, the injection of previous scale information is not straightforward in single-scale DETRs with dense attention. That is, in dense cross-attention only the box center is informed as the reference point, leaving the width and height of the previous box unknown to the current stage, which impedes the current box prediction on basis of the previous one. To better exploit the scale information, DAB-DETR proposes Width \& Height (WH)-Modulated Attention, which modulates the 
Gaussian-like attentional contour to a content-adaptive oval-like shape. However, we find it is sub-optimal due to the 
insufficient use of scale information (Section~\ref{subsec:whm}). 


We 
argue that 
\textit{the cross-attention should explicitly and sufficiently exploit the previous box information before searching for the next positions, rather than using only the box center}. 
Considering the fact that each head in cross-attention separately attends to a small spot and the 
position embedding and inner product are easy 
to highlight a single point 
instead of a box, we present Box Agent, a formulation that condenses the box into head-specific agent points, 
allowing each of them to walk 
around the full scope of the previous box, 
before the cross-attention. By controlling the box agent with 
the decoder embedding, 
the starting reference point can be 
shifted ahead under the guidance of the previous search. 
For instance in the attention head specific to the left side of the box, the previous box center (Fig.~\ref{fig:T_comparison}(a)) is moved to the previous left side of the box (Fig.~\ref{fig:T_comparison}(c)) as the new reference point, in the case that the true left side is outside of the previous box; or the previous box center (Fig.~\ref{fig:T_comparison}(b)) can be directly moved to the left side of the object box (Fig.~\ref{fig:T_comparison}(d)), in the case that the true left side is inside of the previous box. This relieves much burden of the conditional cross-attention.
\begin{figure}[!t]
	\centering
	\includegraphics[width=\linewidth]{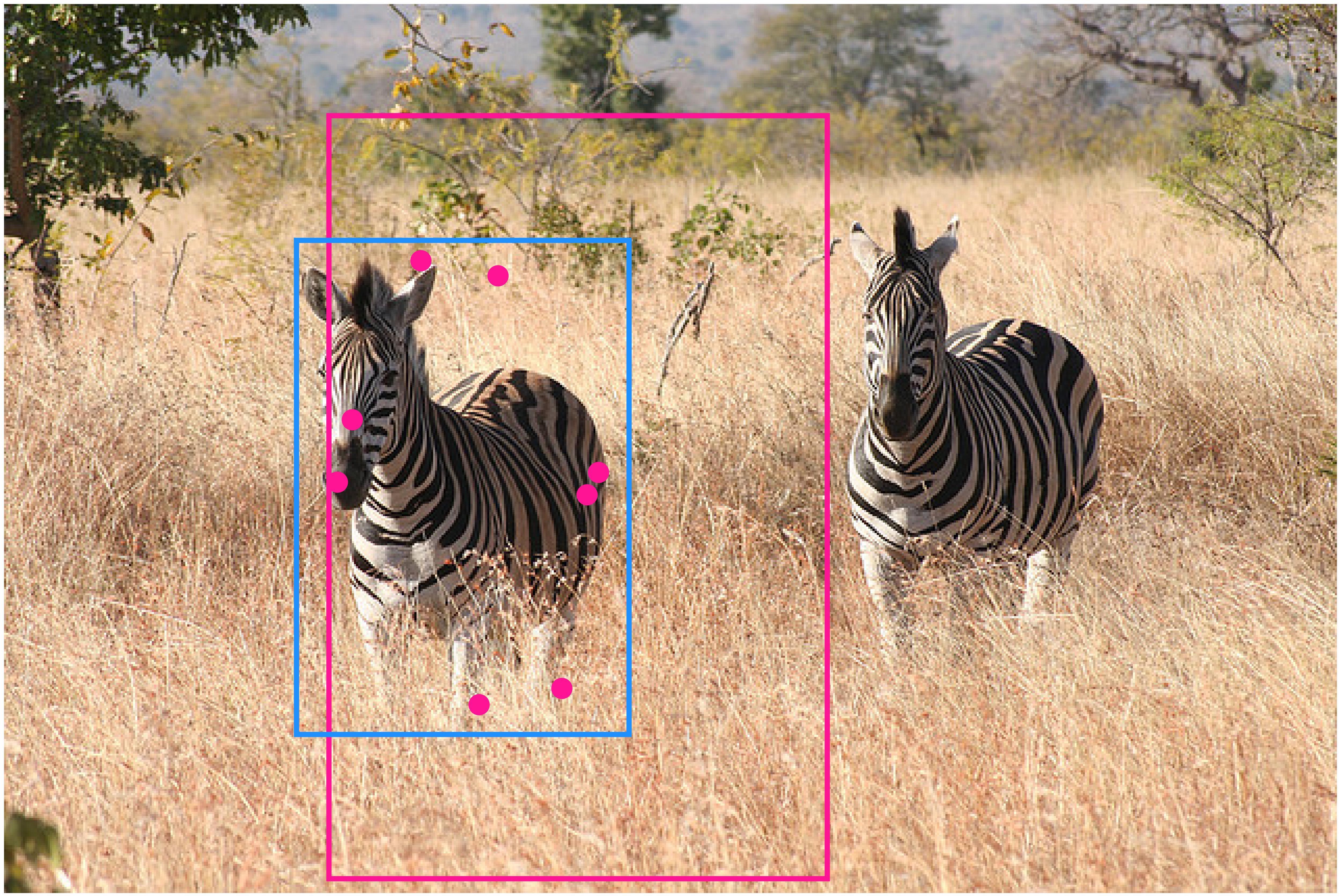}
	\caption{\textbf{How the agent points of the previous box inform current box prediction.} The $8$ head-specific agent points of the previous box (pink)
    jointly contribute to the detection of the four extremities of the current box (blue).
	}
	\label{fig:agents}
\end{figure}
By taking all attention heads into account, all 
agent points jointly contribute to the four extremities of the object box (Fig.~\ref{fig:agents}).


Our formulation directly 
infuses the object scale information into the conditional spatial query. By replacing the 
WH-Modulated Attention in DAB-DETR with Box Agent, 
our single-scale model with ResNet-50 
achieves $44.2$ AP and also faster convergence, 
which suggests \textit{the way for using the object scale information outweighs simply including it}.
In particular, our 
Box Agent only 
requires minor modifications to the code and negligible additional parameters and computational overhead. 
We also conduct 
extensive ablation studies to 
justify that the 
performance gain indeed originates from the 
effective use of the scale information.

\section{Related Work}
\label{sec:related}
We review DETR~\cite{carion2020end}, its variants, and conditional spatial queries in DETR.

\subsection{DETR and Its Variants}
DETR presents an end-to-end object detection model based on Transformer~\cite{vaswani2017attention}, which consists of a CNN backbone used to extract image features, a transformer encoder used to refine the features, and a transformer decoder used to predict the position and object class. 
However, DETR suffers from two main problems. 

The first problem is the slow training convergence.
It requires $500$ training epochs to 
achieve performance comparable to modern detectors~\cite{ren2015faster,he2017mask,cai2018cascade,lin2017focal,tian2019fcos,redmon2018yolov3,law2018cornernet,zhou2019objects,yang2019reppoints,zhang2019bridging,DynamicRCNN}.
Much follow-up work~\cite{zhu2020deformable, gao2021fast, meng2021conditional} solves this problem by improving the cross-attention module in the transformer decoder, and particularly, by associating the query with specific spatial positions of interest. 
Another group of work focuses on query design. Efficient DETR~\cite{yao2021efficient} improves DETR with dense priors at the stage of query initialization. Anchor DETR~\cite{wang2022anchor} directly treats $2$D reference points as queries, while DAB-DETR further studies the role of queries in DETR and 
proposes to use $4$D anchor boxes as queries. Different from these model-level improvements, 
DN-DETR~\cite{li2022dn} introduces query denoising training to alleviate the instability of bipartite graph matching, which is futher improved by DINO~\cite{zhang2022dino}. Also, H-DETR~\cite{jia2022detrs} introduces hybrid matching to improve the one-to-one matching strategy.

The second problem is the high computational complexity. The huge memory footprint and large computational workload of DETR make it not only consume much computation resource but also difficult to leverage multi-scale features. One common solution is the use of sparse attention. PnP-DETR~\cite{wang2021pnp} 
points out the redundancy in 
self-attention and 
extracts part of the features before executing sparse attention. 
Deformable DETR~\cite{zhu2020deformable} transforms dense attention to deformable sparse attention, enabling the use of multi-scale features to improve the performance, but the increased encoder tokens still bring large computational cost. Based on Deformable DETR, Sparse DETR~\cite{roh2021sparse} proposes to update the only required tokens by the decoder, and the additional sparse attention further improves the efficiency of DETR. Lite-DETR~\cite{li2023lite} provides a interleaved way for updating the feature in the transformer encoder to reduce the computational cost.

Our work follows the 
vein that explores the potential of dense cross-attention in the decoder. By leveraging the scale information of the box, we not only accelerate the convergence of DETR but also improve the detection performance.

\subsection{
Conditional Spatial Queries in DETR
}

Under the basis of dense attention in the original DETR~\cite{carion2020end}, Conditional DETR~\cite{meng2021conditional} solves the slow convergence problem of DETR with conditional attention. 
The slow convergence is ascribed to the 
deficiency of high-quality spatial queries at decoder stages, and a conditional spatial query is thus designed conditioned on the decoder embedding and the reference point. In particular, the reference point is first embedded to the same form as the position embedding of the key, 
so that their inner product can mark the position of the reference point as a Gaussian-like shape in the $2$D space. Then a linear projection is generated by the decoder embedding to reweight the position embedding 
of the reference point. The linear projection aims to find positions of interest starting at the reference point for each cross-attention head. Such a design implicitly 
implements the function of generating offsets from the decoder embedding and adding them to the reference point, similar to what is done in Deformable DETR~\cite{zhu2020deformable} and SMCA~\cite{gao2021fast}.

DAB-DETR~\cite{liu2021dab} 
offers a deeper understanding 
on spatial queries. Following Conditional DETR, DAB-DETR 
formulates queries as $4$D dynamic anchor boxes rather than the $256$-dimensional queries and 
updates them layer-by-layer. More importantly, by using anchor box queries, the position information of the box can be explicitly used in the decoder layer, especially the scale information of the box, \ie, width and height. To use the scale information, DAB-DETR proposes WH-Modulated Attention, which 
applies different factors to the $x$ and $y$ 
component of the position embedding to modulate an oval-like, content-aware attention contour. 
However, the reshaping 
contributes little to performance due to 
the insufficient use of scale information.

In this work, we present to use agent points 
to encode the scale information of the box before 
generating the conditional spatial queries.

\section{Understanding Conditional Spatial Queries}
\label{sec:understanding}

We first 
revisit the conditional spatial query in Conditional DETR~\cite{meng2021conditional}, then take a deep look into its role in multi-head mechanism, and finally analyze the 
defect of WH-Modulated Attention in DAB-DETR~\cite{liu2021dab}.

\subsection{Conditional Spatial Query Revisited} 
The authors of Conditional DETR 
ascribe the slow convergence of DETR to the low-quality spatial queries in cross-attention. Hence they 
introduce the conditional spatial query and a conditional cross-attention mechanism, where the cross-attention weight takes the form
\begin{equation}
    \label{eq:cat_cross_attention}
    \mathbf{c}_q^T\mathbf{c}_k+\mathbf{p}_q^T\mathbf{p}_k \,,
\end{equation}
where $\mathbf{c}_q$, $\mathbf{c}_k$, $\mathbf{p}_q$, and $\mathbf{p}_k$ indicate the content query, content key, spatial query, and spatial key, respectively. The addition of $\mathbf{p}_q^T\mathbf{p}_k$ into attention weights are 
used as a soft mask 
to force the content query to 
focus on positions of interest, \eg, the four extremities of the box or some salient object parts. 
Among them, $\mathbf{p}_q$ is the conditional spatial query 
generated by the decoder embedding $\mathbf{f}$ and a reference point $\mathbf{s}$. 
Concretely, $\mathbf{s}$ is first mapped 
to the same form as the position embedding of key with sinusoidal position embedding
\begin{equation}
    \label{eq:sinusoidal}
    \mathbf{p}_s={\tt sinusoidal}(\mathbf{s})\,,
\end{equation}
so that the inner product $\mathbf{p}_s^T\mathbf{p}_k$ can 
highlight the position of the reference point as a Gaussian-like contour on the $2$D space, as shown in Fig.~\ref{fig:T_comparison}(a).
Then a linear projection $\mathbf{T}$ is 
generated from the decoder embedding with a feed-forward network (FFN) as $\mathbf{T}={\rm FFN}(\mathbf{f})$. The projection matrix $\mathbf{T}$ is chosen as a diagonal matrix. Let $\boldsymbol{\lambda}_q$ denote the vector used to construct the diagonal elements in $\mathbf{T}$, then the conditional spatial query is computed using $\boldsymbol{\lambda}_q$ and $\mathbf{p}_s$ by
\begin{equation}
    \label{eq:conditional_spatial_query}
    \mathbf{p}_q=\mathbf{T}\mathbf{p}_s=\boldsymbol{\lambda}_q\odot \mathbf{p}_s\,,
\end{equation}
where $\odot$ is the element-wise multiplication.

Following the same vein, DAB-DETR 
uses dynamic anchor boxes as queries, enabling the layer-by-layer update of boxes, 
and the reference point $\mathbf{s}$ 
is set to be the box center predicted by the previous decoder stage.

\subsection{
Conditional Spatial Queries in Heads
}
\label{subsec:job}

\begin{figure}[!t]
	\centering
	\includegraphics[width=\linewidth]{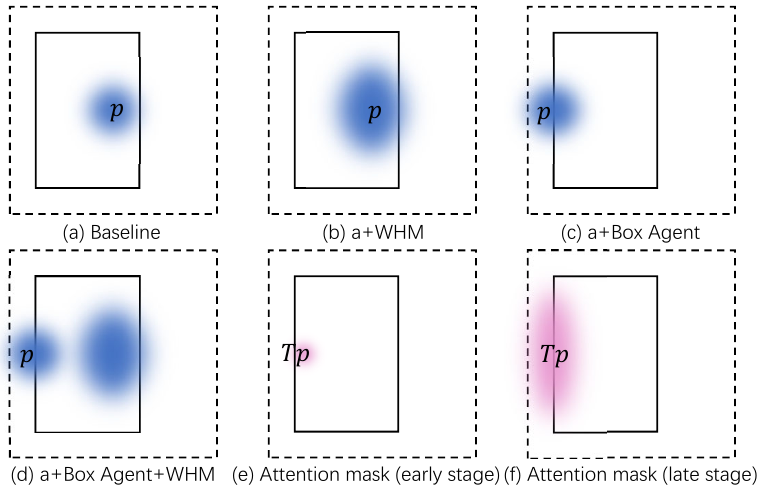}
	\caption{\textbf{The ways of encoding box scale into the attention mask between WH-Modulated Attention (WHM) and Box Agent.} The previous and current box are marked with dotted and solid lines respectively. In (a)-(d) we plot the mask produced by $\mathbf{p}_s^T\mathbf{p}_k$, while the desired mask $(\mathbf{T}\mathbf{p}_s)^T\mathbf{p}_k$ is drawn in (e) and (f). We set DAB-DETR without WHM as the baseline (a), and exhibit the effect of WHM and Box Agent. WHM generates oval attention mask modulated by the box width and height to reflect the box scale, while Box Agent directly moves the reference point in the previous box to the target position.
	}
	\label{fig:whm_comparison}
\end{figure}

We observe that the spatial attention mask produced by $(\mathbf{T}\mathbf{p}_s)^T\mathbf{p}_k$ marks a small point in each head at early stage like in Fig.~\ref{fig:whm_comparison} (e). As the stages go deeper, it hardly changes the position but learns to expand and change the attention shape towards (f). We also visualize the situation at early stage in Fig.~\ref{fig:vis_attn} (stage 2), and that at the final stage can be referred to~\cite{meng2021conditional}. As a result, we decompose the process of the conditional spatial query into two jobs: i) searching for positions of interest starting from a given reference point and ii) expand and change the attention shape to cover representative feature points. And our work focuses on the first job.

Here we have a close look at the multi-head mechanism in DAB-DETR to clarify the first job. 
According to Conditional DETR~\cite{meng2021conditional}, each head in cross-attention focuses on a specific spatial position, 
\eg, one of the four extremities or a certain salient object part. 
As 
mentioned above, the inner product $\mathbf{p}_s^T\mathbf{p}_k$ 
indicates a Gaussian-like 
contour centered at $\mathbf{s}$. 
Then in each cross-attention head, the linear projection $\mathbf{T}$ 
functions to move the attention center to a position of interest, 
say, the left side of an object, by reweighting the elements in $\mathbf{p}_s$, as shown in Fig.~\ref{fig:T_comparison}(a), from blue to pink. 
If the box center is considered to be the reference point as in DAB-DETR, then at every beginning of cross-attention 
of a 
decoder stage $\mathbf{T}$ needs to explore from the previous box center 
to a position of interest for each head. 
Despite the width and height of the box are available in the 
progressively updated 
anchor box queries, they are not 
used in the conditional cross-attention. 
If the whole information of the previous box rather than only its center is informed during the current 
cross-attention, $\mathbf{T}$ can search for positions of interest more effectively. 
In this way, it relieves much burden for $\mathbf{T}$ in the first job.
For instance, when searching for the left side of the object, $\mathbf{T}$ may start from the left side of the previously predicted box, as shown in Fig.~\ref{fig:T_comparison}(c).

\subsection{WH-Modulated Attention in DAB-DETR}
\label{subsec:whm}

\begin{table}[!t]
\centering
\renewcommand{\arraystretch}{1.1}
\addtolength{\tabcolsep}{0pt}
\begin{tabular}{@{}lccc@{}}
\toprule
No. & $wh$-Modulated & $wh$ & ${\rm AP}$ \\
\midrule
1 & & & 42.5\\
2 & \checkmark &  & 42.9\\
3 & \checkmark & \checkmark & 42.8\\
\bottomrule
\end{tabular}
\caption{Study on the effect of object scale information in WH-Modulated Attention.}
\label{tab:ablation_whm}
\end{table}

To 
exploit the scale information 
at the previous stage, DAB-DETR 
introduces WH-Modulated Attention to modulate the attention contour 
during cross-attention. Specifically, it generates reference width and height from the decoder embedding and 
uses the quotient between the reference width/height and the previous box width/height as factors to modulate the attention process, defined by
\begin{equation}
\begin{aligned}
    &{\rm ModulateAttn}\left((x,y),(x_{{\rm ref}},y_{{\rm ref}})\right)=\\
    &\left({\rm PE}(x){\rm PE}(x_{{\rm ref}})\frac{w_{q,{\rm ref}}}{w_q}+{\rm PE}(y){\rm PE}(y_{{\rm ref}})\frac{h_{q,{\rm ref}}}{h_q}\right)/\sqrt{D}\,,
    \nonumber
\end{aligned}
\end{equation}
where $w_q$ and $h_q$ are width and height of the previous box, and $w_{q,{\rm ref}}$ and $ h_{q,{\rm ref}}$ are generated from the decoder embedding. 
With different factors on the $x$ and the $y$ part, it modulates the Gaussian-like contour to an oval-like shape.

To study whether the gain 
of WH-Modulated Attention 
comes from the addition of the 
width and height information of the previous box, we remove $w_q$ and $h_q$ and set 
\begin{equation}
\begin{aligned}
    &{\rm ModulateAttn}\left(((x,y),(x_{{\rm ref}},y_{{\rm ref}})\right)=\\
    &2\left({\rm PE}(x){\rm PE}(x_{{\rm ref}})w_{q,{\rm ref}}+{\rm PE}(y){\rm PE}(y_{{\rm ref}})h_{q,{\rm ref}}\right)/\sqrt{D}
    \nonumber
\end{aligned}
\end{equation}
for a comparison, where the factor $2$ is used to compensate the amplitude loss, because $w_q$ and $h_q$ are typically smaller than $1$. We also 
validate DAB-DETR without WH-Modulated Attention. 
Results are shown in Table~\ref{tab:ablation_whm}. 
We can see that the improvement brought by
WH-Modulated Attention 
is marginal, 
and the use of the width and height information has also little effect in this mechanism.

\section{
Boxing Conditional Spatial Queries
}
\label{sec:method}

After probing the weakness in DAB-DETR, here we 
present our 
solution, Box Agent, which 
condenses the box into head-specific agent points. 
We then provide a perceptual 
understanding of 
the box agent with visualizations. At last, we have a comparison with other similar work.

\begin{figure}[!t]
	\centering
	\includegraphics[width=\linewidth]{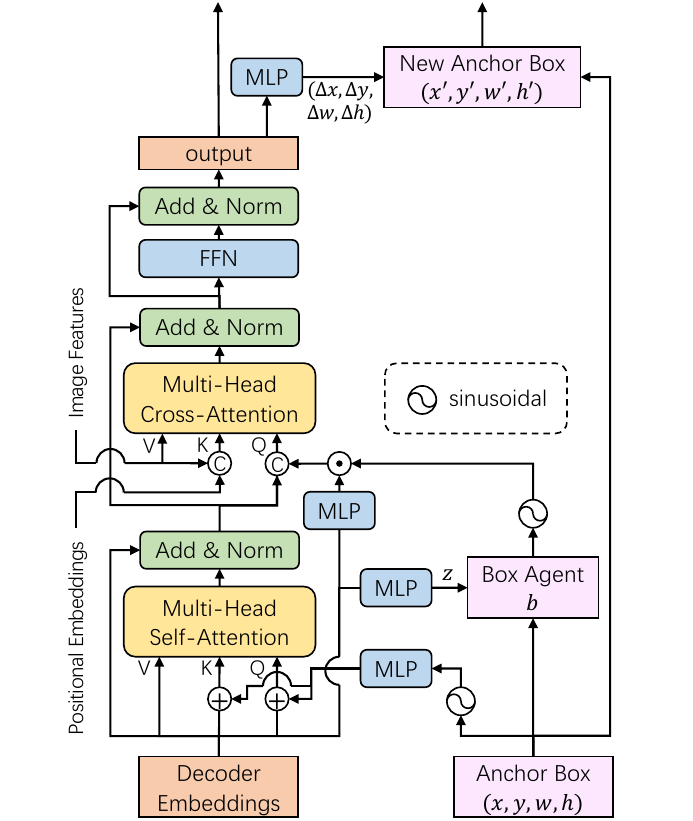}
	\caption{\textbf{One decoder stage in Box-DETR.} For readability, we omit the sigmoid normalization on the anchor boxes. An MLP generates the walker $\mathbf{z}$ from the decoder embedding, which is used to form the box agent $\mathbf{b}$ 
    given the anchor box. Then the conditional spatial query is generated by the position embedded $\mathbf{b}$ and the conditional linear projection, 
    akin to Conditional DETR.
	}
	\label{fig:module}
\end{figure}

\begin{figure*}[!t]
	\centering
	\includegraphics[width=\linewidth]{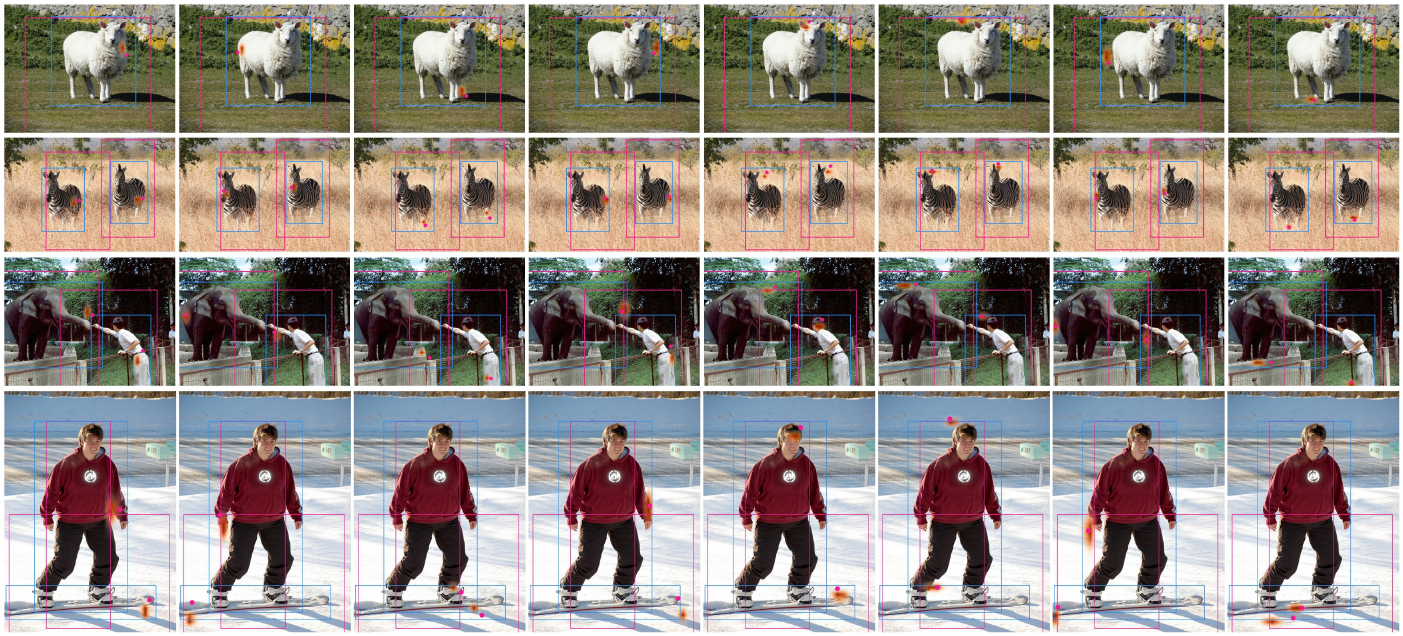}
	\caption{\textbf{Attention visualizations among the $8$ heads in the second decoder layer.} The previous (the first stage) and current (the second stage) box is colored with pink and blue, respectively. The pink point indicates the agent point of the previous box. The spatial attention weights are colored with orange, and the darker color means larger weights.
	}
	\label{fig:vis_attn}
\end{figure*}

\subsection{ 
Box Agent: Representing a Box as Points
}
Given the progressively updating nature of conditional spatial queries,
a reasonable solution may be to determine the starting point used by 
$\mathbf{T}$ under the guidance of the previous search, for instance, somewhere close to the position of interest, as shown in Figures~\ref{fig:T_comparison}(c) and~\ref{fig:T_comparison}(d).
Therefore we need to pass the whole previous box information to the current cross-attention. 
However, the inner product 
between the spatial query and the 
position embedding of key is difficult to characterize a box but is easy to highlight a point. 
Hence our idea is to 
condense the box into an agent point for each cross-attention head. 
We elaborate our design as follows.

First, we simply define the box agent $\mathbf{b}$ by 
\begin{equation}
    \label{eq:bbox}
    \mathbf{b}=({c}_{x},{c}_{y})+({z}_{x}\frac{{w}}{2},{z}_{y}\frac{{h}}{2})\,,
\end{equation}
where ${c}_x$, ${c}_y$, ${w}$, and ${h}$ are the $x$, $y$ coordinate of the box center, box width, and box height, respectively. ${z}_x$ and ${z}_y$ 
are two variables of arbitrary values in range of $[-1,1]$. Eq.~\eqref{eq:bbox} is a common practice to constrain the point in the box. When ${z}_x$ and ${z}_y$ travel all over $[-1,1]$, then $\mathbf{b}$ travels all over the box. This is why we call $\mathbf{b}$ the box agent. Taking multiple agent points into account, they form an equivalent representation of the box.


Again considering the $i$-th head to find the left side of the box, we show the decomposition at Section~\ref{subsec:job} in Fig.~\ref{fig:T_comparison} (c) (d): 1) moving the previous box center to some somewhere close to the object left side; and 2) using $\mathbf{T}$ to highlight the positions of interest more precisely. For instance, $\mathbf{b}_i$ can walk to the previous left side first, by setting $({z}_x, {z}_y)$ as $(-1,0)$ according to Eq.~\eqref{eq:bbox}. In this way we pass not only the center coordinates but the width and height of the previous 
box. The reference point is thus placed at a more reasonable position ahead before used by $\mathbf{T}$.

In Eq.~\eqref{eq:bbox}, we 
generate $\mathbf{z}=({z}_x, {z}_y)$ 
conditioned on the previous search to satisfy the requirement above. In particular, we use the decoder embedding as the controller of $\mathbf{z}$, which 
computes customized $\mathbf{b}_i$ for 
the $i$-th head, thus 
assisting the cross-attention to better localize the positions. Mathematically, $\mathbf{z}={\rm FFN}(\mathbf{f})$ and has the 
size $n\times 2$ in the last dimension, \ie, $z_x$ and $z_y$ for $n$ heads. Then the conditional spatial query $\mathbf{p}_q^i$ specific to $i$-th head has the form
\begin{equation}
    \label{eq:new_conditional_spatial_query}
    \mathbf{p}_{q}^i=\mathbf{T}\mathbf{p}_{b_i}=\boldsymbol{\lambda}_q\odot \mathbf{p}_{b_i}\,,i=1,2,...,n \,,
\end{equation}
where 
$\mathbf{T}$ can either be shared or unshared among heads, which has little influence on performance. We also study whether to explicitly 
constrain $\mathbf{z}$ to $[-1,1]$ or not, which will be 
discussed in the ablation study. 

By replacing the WH-Modulated Attention in DAB-DETR with our Box Agent, we arrive at our new DETR model named Box-DETR. The decoding block 
of Box-DETR is very simple, which is illustrated in Fig.~\ref{fig:module}.


\begin{table*}[!t]
    \centering
    \renewcommand{\arraystretch}{1.1}
    \addtolength{\tabcolsep}{-2.5pt}
    \begin{tabular}{@{}lccccccccccc@{}}
    \toprule
        Model & backbone & multi-scale & \#epochs & ${\rm AP}$ & ${\rm AP}_{50}$ & ${\rm AP}_{75}$ & ${\rm AP}_{S}$ & ${\rm AP}_{M}$ & ${\rm AP}_{L}$ & GFLOPs & Params\\
        \midrule
        DETR~\cite{carion2020end} & R50 &  & 500 & 42.0 & 62.4 & 44.2 & 20.5 & 45.8 & 61.1 & 86 & 41M \\
        Anchor DETR$^*$~\cite{wang2022anchor} & R50 &  & 50 & 42.1 & 63.1 & 44.9 & 22.3 & 46.2 & 60.0 & - & 39M \\
        SMCA~\cite{gao2021fast} & R50 &  & 50 & 41.0 & - & - & 21.9 & 44.3 & 59.1 & 86 & 42M \\
        SAM-DETR-w/SMCA~\cite{zhang2022accelerating} & R50 &  & 50 & 41.8 & 63.2 & 43.9 & 22.1 & 45.9 & 60.9 & 100 & 58M \\
        REGO-DETR~\cite{chen2022recurrent} & R50 &  & 50 & 42.3 & 60.5 & 46.2 & \textbf{26.2} & 44.8 & 57.5 & 112 & 58M \\
        Deformable DETR~\cite{zhu2020deformable} & R50 &  & 50 & 39.0 & 60.2 & 40.7 & 18.2 & 42.9 & 57.9 & - & 35M \\
        Conditional DETR~\cite{meng2021conditional} & R50 &  & 50 & 40.9 & 61.8 & 43.3 & 20.8 & 44.6 & 59.2 & 90 & 44M \\
        DAB-DETR~\cite{liu2021dab} & R50 &  & 50 & 42.8 & 63.3 & 45.7 & 23.2 & 46.2 & 61.4 & 94 & 44M \\
        SAP-DETR~\cite{liu2023sap} & R50 &  & 50 & 43.1 & 63.8 & 45.4 & 22.9 & 47.1 & 62.1 & 92 & 47M \\
        Box-DETR & R50 &  & 50 & 43.8 & 64.0 & 46.5 & 23.4 & 47.9 & 61.5 & 94 & 44M \\
        Box-DETR$^*$ & R50 &  & 50 & \textbf{44.2} & \textbf{64.5} & \textbf{46.8} & 24.0 & \textbf{48.3} & \textbf{62.2} & 100 & 44M \\
        \rowcolor{WhiteSmoke} DN-DAB-DETR~\cite{li2022dn} & R50 &  & 50 & 44.1 & 64.4 & 46.7 & 22.9 & 48.0 & 63.4 & 94 & 44M \\
        \rowcolor{WhiteSmoke} DN-Box-DETR & R50 &  & 50 & \textbf{45.3} & \textbf{65.2} & \textbf{48.5} & \textbf{24.8} & \textbf{48.6} & \textbf{64.5} & 94 & 44M \\
        \rowcolor{WhiteSmoke} DINO-DAB-DETR~\cite{zhang2022dino} & R50 &  & 12 & 40.8 & 60.5 & 43.4 & 20.7 & 44.3 & 58.5 & - & - \\
        \rowcolor{WhiteSmoke} DINO-Box-DETR & R50 &  & 12 & \textbf{41.8} & \textbf{61.4} & \textbf{44.3} & \textbf{21.3} & \textbf{45.2} & \textbf{60.5} & - & - \\
        \midrule
        Faster RCNN-FPN~\cite{ren2015faster} & R50 & \checkmark & 108 & 42.0 & 62.1 & 45.5 & 26.6 & 45.5 & 53.4 & 180 & 42M \\
        SMCA~\cite{gao2021fast} & R50 & \checkmark & 50 & 43.7 & 63.6 & 47.2 & 24.2 & 47.0 & 60.4 & 152 & 42M \\
        Deformable DETR~\cite{zhu2020deformable} & R50 &  \checkmark & 50 & 46.2 & 65.0 & 50.0 & 28.3 & 49.2 & 61.5 & 173 & 40M \\
        \midrule
        DETR~\cite{carion2020end} & R101 &  & 500 & 43.5 & 63.8 & 46.4 & 21.9 & 48.0 & 61.8 & 152 & 60M \\
        Anchor DETR$^*$~\cite{wang2022anchor} & R101 &  & 50 & 43.5 & 64.3 & 46.6 & 23.2 & 47.7 & 61.4 & - & 58M \\
        Conditional DETR~\cite{meng2021conditional} & R101 &  & 50 & 42.8 & 63.7 & 46.0 & 21.7 & 46.6 & 60.9 & 156 & 63M \\
        DAB-DETR~\cite{liu2021dab} & R101 &  & 50 & 43.7 & 64.3 & 46.9 & 23.9 & 47.7 & 60.9 & 174 & 63M \\
        SAP-DETR~\cite{liu2023sap} & R101 &  & 50 & 44.4 & 64.9 & 47.1 & 24.1 & 48.7 & 63.1 & 158 & 67M \\
        Box-DETR & R101 &  & 50 & 44.6 & 64.6 & 47.7 & 24.8 & 48.6 & 62.4 & 174 & 63M \\
        Box-DETR$^*$ & R101 &  & 50 & \textbf{45.3} & \textbf{65.3} & \textbf{48.9} & \textbf{26.0} & \textbf{49.6} & \textbf{63.5} & 179 & 63M \\
        \midrule
        Faster RCNN-FPN~\cite{ren2015faster} & R101 & \checkmark & 108 & 44.0 & 63.9 & 47.8 & 27.2 & 48.1 & 56.0 & 246 & 60M \\
        SMCA~\cite{gao2021fast} & R101 & \checkmark & 50 & 44.4 & 65.2 & 48.0 & 24.3 & 48.5 & 61.0 & 218 & 50M \\
        \midrule
        \rowcolor{WhiteSmoke} DN-Box-DETR & Swin-B &  & 36 & 51.0 & 72.1 & 54.4 & 29.9 & 56.0 & 72.5 & - & 107M\\
        \rowcolor{WhiteSmoke} DN-DAB-DETR~\cite{li2022dn} & Swin-L &  & 36 & 51.5 & 73.1 & 55.3 & 29.1 & 56.9 & \textbf{73.2} & - & 215M\\
        \rowcolor{WhiteSmoke} DN-Box-DETR & Swin-L &  & 36 & \textbf{52.0} & \textbf{73.3} & \textbf{55.7} & \textbf{30.5} & \textbf{57.0} & 73.0 & - & 215M\\
    \bottomrule
    \end{tabular}
    \caption{Comparison with other detectors (mainly single-scale DETRs). Except that DETR uses $100$ queries, all other DETR variants use $300$.
    * indicates using $3$ pattern embeddings as in Anchor DETR~\cite{wang2022anchor}. 
    Deformable DETR is tested with the `+iterative bbox refinement' version for fair comparison. Best performance is in \textbf{boldface}.}
    \label{tab:main_result}
\end{table*}

\subsection{Visualization of Agent Points}

Here we visualize the spatial attention maps 
computed by $(\mathbf{T}\mathbf{p}_{b_i})^T\mathbf{p}_k$, the currently predicted box, the previous box, and 
the agent points of the previous box in cross-attention at the second decoder stage in Fig.~\ref{fig:vis_attn}. The previous box from the first decoder layer and its corresponding agent point in a certain head is marked in pink, while the current box is marked as blue. The attention amplitude is encoded in orange, and the darker color means larger weights. We can 
see that the current positions of interest are 
closely related to the agent points of the previous box, and the agent points search for the extremities of the box and some salient object parts. For example, at the row 2 column 4 in Fig.~\ref{fig:vis_attn}, 
the previous boxes are condensed into the points at the right side of the zebras, and the attention weights are also gathered here. 
All the $8$ agent points 
jointly contribute to the location of 
the $4$ extremities of the object box (see also Fig.~\ref{fig:agents}).

\begin{figure}[!t]
	\centering
	\includegraphics[width=\linewidth]{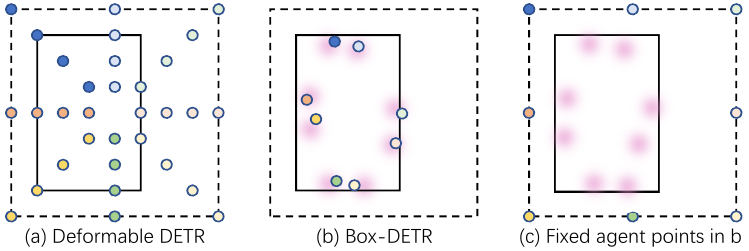}
	\caption{\textbf{Comparison between Deformable DETR and Box-DETR.} The previous and current box are marked with dotted and solid lines respectively. The circles indicate the initial sampling positions in Deformable DETR in (a), while denote agent points in Box-DETR in (b) and (c). Different colors indicate different heads. The sparse cross-attention in Deformable DETR evenly sets initial sampling positions in the previous box to encode the scale information. Box-DETR uses content-aware agent points to guide the dense cross-attention. If one sets fixed reference points in dense attention, since each head corresponds to one reference point, the total $8$ points cannot cover enough space to capture the current box. We draw one of this scenarios in (c), where the projection $\mathbf{T}$ still struggles to attend to the correct position. 
	}
	\label{fig:deform_comparison}
\end{figure}

\begin{figure}[!t]
	\centering
	\includegraphics[width=\linewidth]{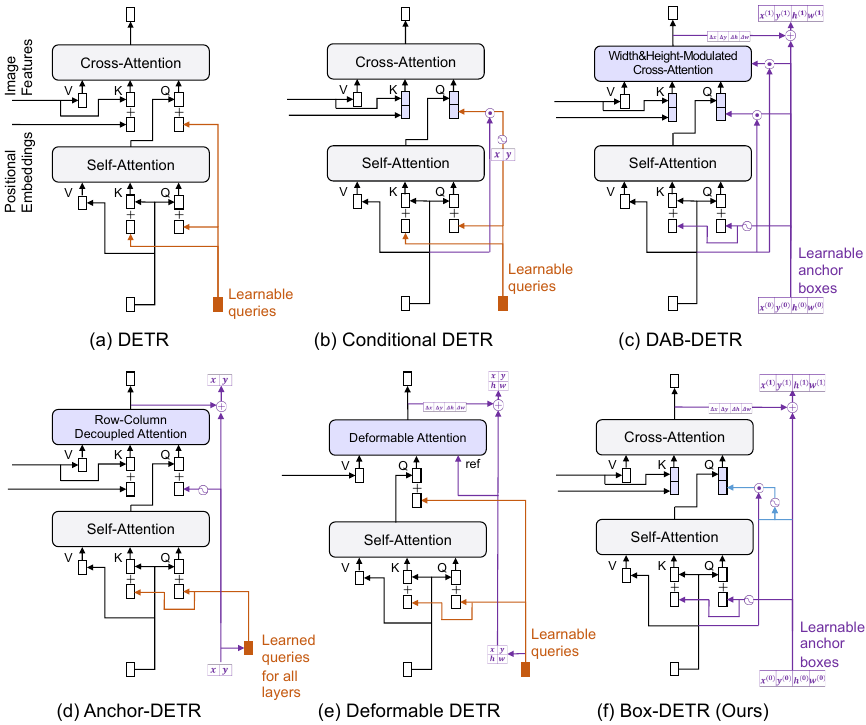}
	\caption{\textbf{Comparison of the decoder module between Box-DETR and other DETR variants.} Based on DAB-DETR, the decoder of Box-DETR removes the WH-Modulated Attention and generates head specific agent points to form the spatial query.
	}
	\label{fig:comparison_related}
\end{figure}

\paragraph{Comparison between WH-Modulated Attention and Box Agent.}
As shown in Fig.~\ref{fig:whm_comparison}, (a) we set DAB-DETR without WHM as the baseline and (e) (f) plot the desired attention mask. (b) WHM does not change the reference point, and generates an oval attention mask.
(c) Box Agent changes reference (agent) points for each head. In cross-attention, each head separately attends to a small spot, instead of a box region. Therefore the point-level mask seems more suitable than the box-level one. (d) With WHM, Box Agent is disturbed. This also explains the ablation study about their complementarity in Table~\ref{tab:ablation_complementary}. 

\paragraph{Comparison between Deformable DETR and Box-DETR.}
In the `+iterative bbox refinement' version of Deformable DETR~\cite{zhu2020deformable}, there is a similar process to Eq.~\eqref{eq:bbox} for the offset generation. However, the way to encode the box scale information varies. As shown in Fig.~\ref{fig:deform_comparison}, Deformable DETR evenly sets grid-shape initial sampling positions in the previous box, which can be seen as the reference points in dense attention. And then it generates offsets around the initial positions, which is similar to the function of the projection matrix $\mathbf{T}$. Though the initialization for sparse attention in Deformable DETR encodes the previous scale information, it can not be used in single-scale DETRs with dense attention. Since in dense attention DETRs the number of reference points are up to $8$ (equal to the head numbers), the fixed agent points cannot cover enough space to capture the current box.

\textbf{Comparison with other DETR variants in the model level.}
We borrow the sketch map of different DETR variants from DAB-DETR~\cite{liu2021dab}, and add Box-DETR for model comparison, as shown in Fig.~\ref{fig:comparison_related}. The models include: the original DETR~\cite{carion2020end}, Conditional DETR~\cite{meng2021conditional}, DAB-DETR~\cite{liu2021dab}, anchor DETR~\cite{wang2022anchor}, Deformable DETR~\cite{zhu2020deformable} and our Box-DETR. DAB-DETR has carefully compared DAB-DETR and others, and the main difference is that DAB-DETR uses anchor boxes as queries, and updates them layer-by-layer in the case of dense attention. Compared with DAB-DETR, the cross-attention reference points are head-specific agent points in Box-DETR, rather than the head-shared previous box center in DAB-DETR. And we remove the Width \& Height Modulated Attention in DAB-DETR.

\section{Results and Discussion}

In this section, we present and discuss our main experimental results. 

\subsection{Dataset, Metrics, Baselines, and Protocols}
We conduct experiments on the MS-COCO 2017 dataset~\cite{lin2014microsoft}. The models are trained on its train split and evaluated on its 
validation set with the AP metric. We choose DAB-DETR~\cite{liu2021dab} as our baseline and select ResNet-50 (R50), ResNet-101 (R101)~\cite{he2016deep}, Swin-B and Swin-L~\cite{liu2021swin} as the backbones. 
We use the codebase provided by~\cite{liu2021dab} with dropout as $0.0$ instead of $0.1$ described in their paper and also retrain DAB-DETR, which produces slightly better results. 
Inspired by Anchor DETR~\cite{wang2022anchor}, we also test $3$ pattern embeddings postfixed with $*$. For the $1\times$ ($12$ epochs) and the $3\times$ ($36$ epochs) setting, the learning rate drops at the $11$-th and $30$-th epoch by multiplying $0.1$, respectively. For DN-DETR~\cite{li2022dn}, we only replace the DAB-DETR model with Box-DETR, and 
keep other parts unchanged.

\subsection{Main Results}
Since we focus on single-scale DETRs with dense attention, we mainly compare our Box-DETR with single-scale DETR-like models and cite a few multi-scale ones for reference. We also report the performance under different training schedules and with the query denoising~\cite{li2022dn,zhang2022dino} setting. 

\vspace{-5pt}
\paragraph{Comparison With DETR Variants.} 
Quantitative results are shown in Table~\ref{tab:main_result}. Our Box-DETR surpasses the state-of-the-art single-scale model DAB-DETR by $1$ AP, and achieves $44.2$ and $45.3$ AP with the R50 and the R101 backbone, respectively. Compared with DAB-DETR, Box-DETR only costs $3.6$K 
additional parameters. From 
Table~\ref{tab:main_result}, we can also see that 
the gain mainly comes from detecting medium objects for R50, with a significant $1.7$ AP higher than DAB-DETR. However, with R101, the gain of performance originates from detecting objects 
at all scales. It also outperforms the original DETR with $500$ training epochs by large margins ($+2.2$ AP with R50 and $+1.8$ AP with R101). Box-DETR surpasses the strong multi-scale CNN baseline Faster-RCNN by a large margin ($+2.2$ AP with R50 and $+1.3$ AP with R101) with fewer training epochs ($50$ vs. $108$). Box-DETR also outperforms the multi-scale DETR based model SMCA with R50 and R101.

\begin{figure}[!t]
	\centering
	\includegraphics[width=\linewidth]{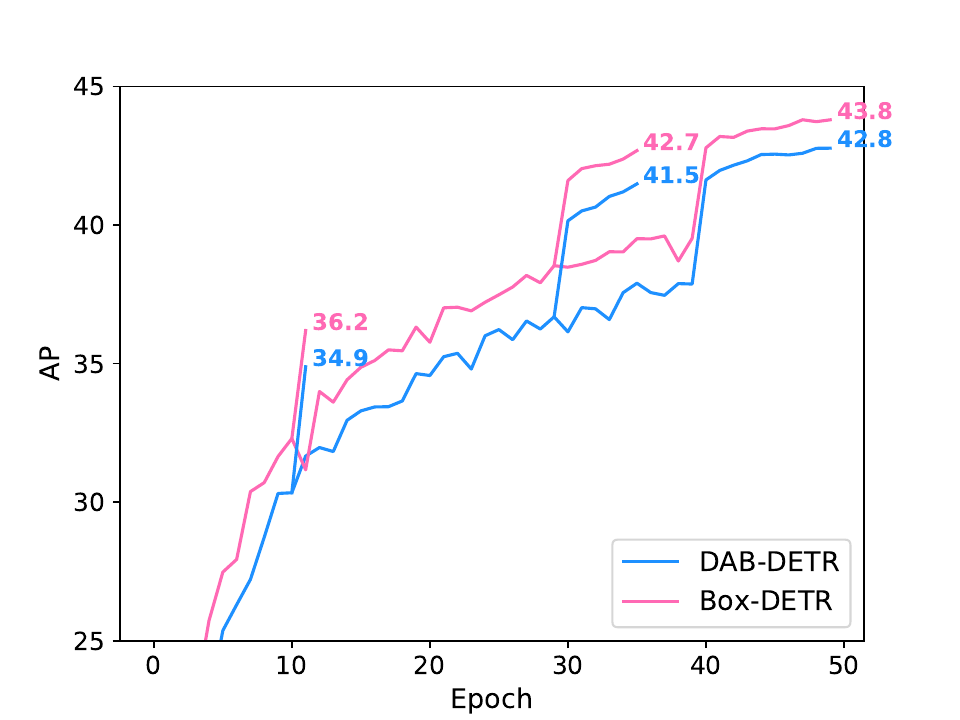}
	\caption{\textbf{Learning curve for DAB-DETR and Box-DETR.} The models are trained with ResNet-50.
	}
	\label{fig:learning_curve}
\end{figure}

\vspace{-5pt}
\paragraph{Comparison With DAB-DETR Under the $1\times$ and $3\times$ Schedules.} 
Here we compare the performance between DAB-DETR and Box-DETR under the standard $1\times$ ($12$ epochs) and $3\times$ ($36$ epochs) settings with R50 backbone. 
The learning curves of DAB-DETR and Box-DETR are shown in Fig.~\ref{fig:learning_curve}. 
We observe that Box-DETR surpasses DAB-DETR by $1.3$ AP under the $1\times$ schedule and $1.2$ AP under the $3\times$ schedule. 


\vspace{-5pt}
\paragraph{Training With Query Denoising.} Here we further train our model under the stronger Query Denoising~\cite{li2022dn,zhang2022dino} setting with R50 backbone. 
Quantitative results are shown in Table~\ref{tab:main_result}. With Box-DETR, DN-Box-DETR outperforms the baseline DN-DAB-DETR by $1.2$ AP with R50 in $50$ epochs, and $0.5$ AP with Swin-L in $36$ epochs. Similarly, DINO-Box-DETR invites $1.0$ AP improvement than DINO-DAB-DETR with R50 in $12$ epochs, which 
proves the effectiveness and the complementation of Box Agent 
to existing techniques. 

\begin{figure*}
\begin{minipage}{\textwidth}

\makeatletter\def\@captype{table}\makeatother
\begin{minipage}{.32\textwidth}
	\centering
    \renewcommand{\arraystretch}{1.1}
    \addtolength{\tabcolsep}{-2pt}
    \begin{tabular}{@{}lcccc@{}}
    \toprule
    No. & Box Agent & $wh$ & $\sigma$-Norm & ${\rm AP}$ \\
    \midrule
    1 & \checkmark  &       &             & 41.0\\
    2 & \checkmark  &       & \checkmark  & 42.5\\
    3 &             &       &             & 42.5\\
    4 & \checkmark  & \checkmark &        & 43.8\\
    
    \bottomrule
    \end{tabular}
    \caption{Ablation study on the object scale information.}
    \label{tab:ablation_scale}
\end{minipage}
\hfill
\makeatletter\def\@captype{table}\makeatother
\begin{minipage}{.28\textwidth}
    \centering
    \renewcommand{\arraystretch}{1.1}
    \addtolength{\tabcolsep}{-3pt}
    \begin{tabular}{@{}lccc@{}}
    \toprule
    No. & Backbone & $\tt tanh$-Norm & ${\rm AP}$\\
    \midrule
    1 & R50 &  & 43.8\\
    2 & R50 & \checkmark & 43.2\\
    3 & R101 &  & 44.6\\
    4 & R101 & \checkmark & 44.5\\
    \bottomrule
    \end{tabular}
    \caption{Ablation study on the effect of normalization on $\mathbf{z}$.}
    \label{tab:ablation_norm}
\end{minipage}
\hfill
\makeatletter\def\@captype{table}\makeatother
\begin{minipage}{.34\textwidth}
    \centering
    \renewcommand{\arraystretch}{1.1}
    \addtolength{\tabcolsep}{-0.5pt}
    \begin{tabular}{@{}lccc@{}}
    \toprule
    No. & $wh$-Modulated & Box Agent & ${\rm AP}$ \\
    \midrule
    1 &  & \checkmark & 43.8\\
    2 & head-shared & \checkmark & 43.0\\
    3 & head-unshared & \checkmark & 42.8\\
    \bottomrule
    \end{tabular}
    \caption{Ablation study on the complementarity between WH-Modulated Attention and Box Agent.}
    \label{tab:ablation_complementary}
\end{minipage}

\end{minipage}
\end{figure*}

\begin{figure}[!t]
	\centering
	\includegraphics[width=\linewidth]{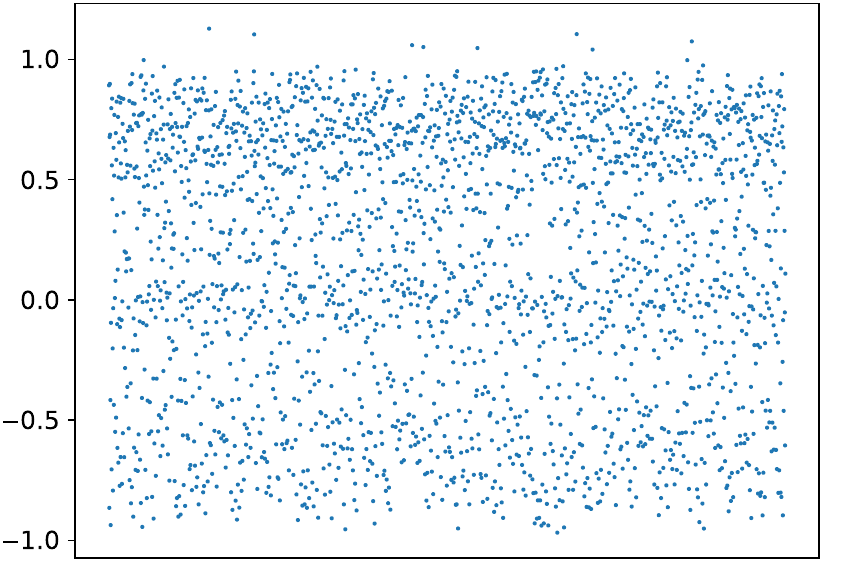}
	\caption{\textbf{Scatter diagram of $z_x$ distribution in the unnormalized setting.} For a randomly chosen testing image, we plot all $300$ (queries) $\times8$ (heads) $z_x$'s at the second decoder stage in a sequential manner (the $x$-axis) for ease of illustration. Most $z_x$'s are in range of $[-1,1]$ even without normalization, while only a tiny minority exceeds the range. 
    We can reasonably infer that when the number of agent points goes to infinity, then $\mathbf{z}$ would cover $[-1,1]^2$ and $\mathbf{b}$ would cover the previous box, which verifies that the Box Agent indeed informs the information of the previous box.
	}
	\label{fig:z_range}
\end{figure}

\subsection{Ablation Study}
To figure out whether the gain of performance originates from the scale information and to justify each design of 
Box Agent, here we conduct 
a number of ablation studies.

\vspace{-5pt}
\paragraph{Effectiveness of Scale Information.}
We 
first justify that the gain of performance comes from the addition of scale information. With Eq.~\eqref{eq:bbox}, we set the decoder embedding $\mathbf{f}$ as the controller of $\mathbf{z}$. A natural 
question may be that perhaps the predicted offset $\mathbf{z}$ 
brings the improvement, while the scaling factors $w$ and $h$ 
contribute little. Hence, we set two 
additional baselines where the scaling factors are removed.
For the first baseline (No. 1), 
we remove $w$ and $h$, and set 
\begin{equation}
    \mathbf{b}=(c_{x},c_{y})+(z_{x},z_{y})\,.
    \nonumber
\end{equation}
For the second baseline (No. 2), we set 
\begin{equation}
    \mathbf{b}=\sigma(\sigma^{-1}((c_{x},c_{y}))+(z_{x},z_{y}))\,,
    \nonumber
\end{equation}
where $\sigma$ is the ${\tt sigmoid}$ function. We add this baseline to constrain all $\mathbf{b}$'s to be in $[0,1]$, 
because the negative $\mathbf{b}$ indicates that a point is outside of the image. We also compare their performance against Box-DETR without (No. 3) and with (No. 4) the box agent and scale information. 

As shown in Table~\ref{tab:ablation_scale}, if we formulate the agent point without the object scale information, it 
even worsens the performance (No. 3 vs.\ No. 1 \& No. 2). In contrast, by infusing the information of the whole box into the agent point, our box agent 
unearths the value of object scale information, leading to a $1.3$ AP improvement. This proves that the gain of performance mainly comes from the object scale information, rather than the predicted offsets.

\vspace{-5pt}
\paragraph{Effect of Normalization on $\mathbf{z}$.}
In our formulation, we do not explicitly constrain the value range of $\mathbf{z}$. Here we see whether such a constraint helps. In particular, we constrain $\mathbf{z}$ to $(-1,1)$ using the ${\tt tanh}$ function and reformulate
\begin{equation}
    \mathbf{b}=(c_{x}\,,c_{y})+({\tt tanh}(z_{x})\frac{w}{2}\,,{\tt tanh}(z_{y})\frac{h}{2})\,.
    \nonumber
\end{equation}
Qualitative results are shown in Table~\ref{tab:ablation_norm}. We can observe that R50 seems 
more sensitive to normalization. Further, during training with R50 the learning curve shows that at the early stage the AP without normalization is lower than that with normalization, but as training continues the box agent without normalization surpasses that with normalization. We 
explain this phenomenon as the trade-off between exploration and exploitation. With normalized $\mathbf{z}$, $\mathbf{b}$ strictly walks around the scope of the previous box as full exploitation, while unnormalized $\mathbf{z}$ allows some deviations from the previous box and can explore more. At the early stage, the model does not learn reliable representation of objects, so the normalized $\mathbf{z}$ that only strictly uses the information of the previous layer behaves better. However, when the model has been trained sufficiently well, the additional exploration with unnormalized $\mathbf{z}$ would be preferred. For example in Fig.~\ref{fig:vis_attn}, for the first, fourth, and seventh subfigure in the last row, in these heads the agent point exceeds the boundary of the previous box but reaches a more reasonable point. However, given a stronger backbone R101, the difference is narrow ($44.5$ vs. $44.6$). A statistical analysis of $\mathbf{z}$ can be referred to Fig.~\ref{fig:z_range}, which suggests the explicit normalization seems unnecessary.

\vspace{-5pt}
\paragraph{Dynamic or fixed agent points.}
We use the decoder embedding produced by the previous stage to generate $\mathbf{z}$, because the search of the current box is conditioned on the previous search. We verify its effectiveness by setting a compared baseline with fixed agent points in the previous box. If the agent points are fixed like in Fig.~\ref{fig:deform_comparison} (c), the AP metric degrades from $43.8$ to $42.7$ (R50, 50 epochs).
\vspace{-5pt}
\paragraph{Complementarity Between WH-Modulated Attention and Box Agent.} We wonder if 
WH-Modulated Attention and our Box Agent can 
work collaboratively. 
Based on Box-DETR, we add WH-Modulated Attention for each head. The results in Table~\ref{tab:ablation_complementary} show that WH-Modulated Attention can have an inverse effect on the performance, no matter in a head-shared or a head-unshared manner, as discussed in Section~\ref{sec:method}.


\section{Conclusion}
In this paper, we first study the function of conditional spatial queries in DAB-DETR, and point out the insufficient use of scale information, for only informing the previous box center to the cross-attention as the reference point. 
Targeting this problem, 
we propose Box Agent. Its key idea is to represents a box as points. It condenses the box into head-specific agent points to encode the scale information. By using them as the new reference points, the cross-attention can learn the location of the object box more effectively. Experimental experiments show that Box Agent contributes to not only accelerated convergence but also improved performance, and we have also carefully justified that 
the improvement 
indeed comes from 
the effective use of the scale information. 

\vspace{-5pt}
\paragraph{Limitations.} The current design of the box agent is only for DETRs with dense attention. Due to the large computational cost of dense attention, it cannot directly applicable to multi-scale detectors. We plan to address this in the future.

\vspace{-5pt}
\paragraph{Acknowledgement.} This work is supported by the National Natural Science Foundation of China Under Grant No. 62106080.

{\small
\bibliographystyle{ieee_fullname}
\bibliography{egbib}
}

\end{document}